> REPLACE THIS LINE WITH YOUR PAPER IDENTIFICATION NUMBER (DOUBLE-CLICK HERE TO EDIT) <    1

# Static and Dynamic Synthesis of Bengali and Devanagari Signatures

Miguel A. Ferrer, Sukalpa Chanda, Moises Diaz, Chayan Kr. Banerjee, Anirban Majumdar, Cristina Carmona-Duarte, Parikshit Acharya, Umapada Pal

*Abstract*—Developing an automatic signature verification system is challenging and demands a large number of training samples. This is why synthetic handwriting generation is an emerging topic in document image analysis. Some handwriting synthesizers use the motor equivalence model, the well-established hypothesis from neuroscience, which analyses how a human being accomplishes movement. Specifically, the motor equivalence model divides human actions into two steps: (i) the effector independent step at cognitive level and (ii) the effector dependent step at motor level. In fact, recent work reports the successful application to Western scripts of a handwriting synthesizer, based on this theory. This paper aims to adapt this scheme for the generation of synthetic signatures in two Indic scripts, Bengali (Bangla) and Devanagari (Hindi). For this purpose, we use two different online and offline databases for both Bengali and Devanagari signatures. Our paper reports an effective synthesizer for static and dynamic signatures written in Devanagari or Bengali scripts. We obtain promising results with artificially generated signatures in terms of appearance and performance when we compare the results with those for real signatures.

## I. INTRODUCTION

The handwritten signature has been used for many centuries for authenticating personal identity. Performing signature verification aided by computer vision systems is nowadays common practice. However, because of large intra-user variability, obtaining reliable performance of such verification systems demands exhaustive system training and requires a large number of samples. Procuring a signature database of real signatures that could be used in extensive training and testing is a tedious, complex and costly task since a reliable database should exhibit biometric variability as much as possible (i.e. multi-session, multiple acquisition sensors, different signal quality, emotions, etc.). User cooperation is also a factor. On top of that, legal issues pertaining to data protection are also relevant.

To address these problems, researchers have investigated methods of generating synthetic signature data that exhibits similar characteristics to real life signatures. However, all such methods used so far focus on Western-based scripts. Present state-of-the art research reveals this fact.

### A. Schemes for Synthetic Signature Generation

Various workers have proposed methods for synthetic generation, based on dynamic signature duplication, aimed at improving enrollment in dynamic Western signature verification ([1][2][3][4]). In [5] researchers generate new static signatures from two dynamic samples produced by the same user. A static handwritten Western signature database is generated in [6] by making an affine transformation of the original signatures. Similar objectives were achieved in [7] by the generation of enhanced static signatures which approximate the performance of real Western signature datasets. Workers in [8] use an inspired cognitive approach to generate Western static versions of real dynamic signatures and in [9] to generate Western static signatures from real static ones. Those in [10] use a similar proposal to generate duplicates of Bengali signatures.

Popel proposes an approach for completely new signature generation in [11]. He develops a language model based on visual characteristics extracted from the time domain. However, authors in [11] provides only visual validation without any performance evaluation. Galbally et al. [12][13] propose a novel methodology for the generation of genuine Western synthetic online signatures using flourish and isolated casual characters. They fused spectral analysis and the kinematic theory of rapid human movement [14][15][16][17] to generate the master signature of a new identity. Similarly, [18] reports a proposal to use heuristic procedures to generate Western static genuine and forged signatures.

Recently, elaborated procedures to generate signatures have been proposed based on the motor equivalence theory. Specifically, in [19] for Western offline signatures and in [20] for Western online and offline signatures simultaneously.

### B. The Motor Equivalence Model

The motor equivalence theory, originally formulated by Lashley [21] and later articulated by Bernstein [22], suggests that the movements aimed at performing a single task is dissected by the brain into two steps. The first step is called "effector-independent". It stores the movement in an abstract form as a spatial sequence of points representing the action

This study was funded by the Spanish government's MIMECO TEC2016-77791-C4-1-R research project and European Union FEDER program/funds.

Miguel a. Ferrer, Moises Díaz and Cristina Carmona are with the Instituto Universitario para el Desarrollo Tecnológico y la Innovación en Comunicaciones, Universidad de Las Palmas de Gran Canaria, Campus de Tafira, Las Palmas de Gran Canaria, Spain. Emails: mferrer@idetic.eu, mdiaz@idetic.eu, ccarmona@idetic.eu

Sukalpa Chanda, Chayan Kr. Banerjee, Anirban Majumdar, Parikshit Acharya and Umapada Pal are with the Computer Vision and Patter Recognition Unit of the Indian Statistical Institute, 203 B.T. Road, 700108 Kolkata, India. umapada@isical.ac.in

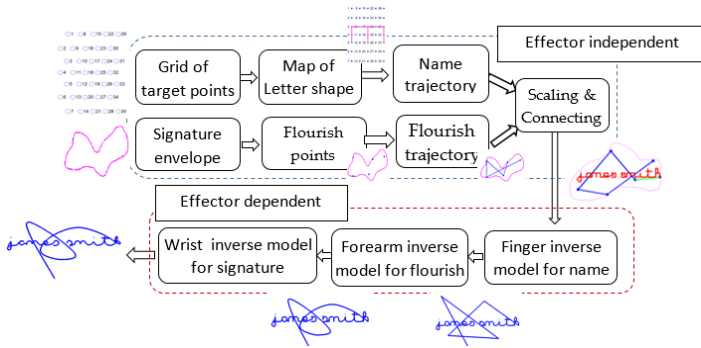

Figure 1. Block diagram of the handwriting signature synthesizer based on the motor equivalence model for Western style.

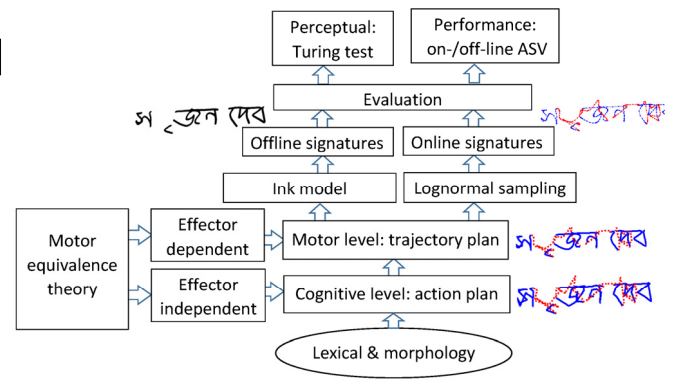

Figure 2. Block diagram of the handwriting signature synthesizer based on the motor equivalence model for Indian script.

plan. The parietal cortex in general, and the posterior parietal cortex and the occipitotemporal junction in particular, are suggested in [23] as the most important brain regions for representation of the action plan. The second step is called "effector dependent". It consists of a sequence of motor commands directed at obtaining particular muscular contractions and articulatory movements in order to execute the action plan [24] .

Applying the motor equivalence model to handwriting, the action plan may be represented by means of strokes, which are encoded in terms of relative positions and spatial directions. Once the movement has been planned, the motor control delivers the commands to specific muscles to produce the handwriting.

In [25] Kawato states that, due to slow biological feedback, the feedback control cannot solely execute fast and coordinated movements. He asserts that the brain acquires an inverse image of the object controlled by motor learning. Accordingly, the brain calculates the motor command by looking at the internal inverse model of the limbs, which are created by the cerebellum. This means that the handwriting of a human being at an early age demands attention while writing and has a clumsy appearance and slow execution speed. However, after prolonged practice, the movements become quick, smooth, automatic and require a minimal cognitive contribution.

### C. Signature Synthesizer based on Motor Equivalence Model

The above factors are combined in the model used for synthetic Western signature generation in [19] and [20] as follows: first, it defines a user grid based on the cognitive map. Second, it represents the name and flourish engrams as a sequence of grid nodes and their stroke limits. To generate these engrams a language model is required. Third, it designs the signature trajectory by applying a motor model to the signature engram. Fourth, from the signature trajectory, it generates the dynamic signature by lognormal sampling of the signature trajectory. Finally, from the same signature trajectory, it generates the static signature image by applying an ink deposition model.

A detailed block diagram of this procedure is shown at Figure 1 for Western signatures and complemented with Figure 2 for Indic signatures.

The above proposed scheme gave us poor results in both appearance and performance when it was applied to synthesize Devanagari and Bengali signatures. On the one hand, the appearance was not perceptually natural for Indian native people and, on the other hand, the performance was significantly different compared to the results obtained with real Indic datasets.

Moreover, a deeper analysis revealed several limitations of the Western-based model proposed in [19][20] to generate Indic scripts. Such limitations are related to the Indic scripts themselves and the handwriting style of Indian writers. Specifically, the Indic scripts consist of more complex characters, shorter strokes and more pen-ups than Western handwriting. Additionally, the Indian writers exhibit more curved pen-up trajectories and pen-up to pen-down transitions without speed minimum, i.e. Western writers lean the pen against the sheet and start to write while Indian writers seem to start the writing in the air. This effect is necessary for rapid handwriting with a high number of pen-up to pen-down transitions. These findings explain how the motor system influences the user variability.

Therefore, a redesign of the signature synthesizer was mandatory to generate Indic handwriting. As a result, this article describes the procedure to generate synthetic Bengali and Devanagari signatures by defining Indic morphology and language models, script shapes and Indic handwriting styles to the synthesizer proposed in [20]. The quality evaluations of the generated Indic signatures are based on both performance and subjective opinion surveys. The performance experiment compares the similarity between the Equal Error Rates (EER) obtained with real and synthetic signatures database with different online and offline classifiers. Instead, the perceptual experiment explores the human confusion between synthetic and real Indic signatures.

The outline of the remainder of the paper is as follows: Section II describes the signature morphology and language model of both Bengali and Devanagari signatures. Section III is devoted to offline Indian signature generation while Section IV is dedicated to the synthesis of online Indian signatures. The generation of multiples samples is discussed in Section V. The experiment and related discussions are described in Section VI and Section VII concludes the article.

> REPLACE THIS LINE WITH YOUR PAPER IDENTIFICATION NUMBER (DOUBLE-CLICK HERE TO EDIT) < 3

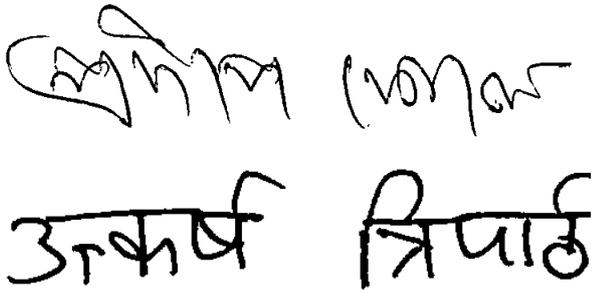

Figure 3. Example of signatures in two different scripts: Bengali (top) and Devanagari (bottom).

## II. MORPHOLOGY AND LANGUAGE MODEL FOR DEVANAGARI AND BENGALI SIGNATURES.

To define an appropriate synthetic signature specimen, a morphology and language model is needed to construct signatures with similar appearance to the original signature [26]. This requirement is mainly due to the fact that performance of a signature database depends on characteristics such as the average number of words, letters per word, etc. Additionally, although real names are avoided for privacy reason, readable names are recommended for perceptual acceptability of the synthetic signature.

### A. Bengali Signature Characteristics

Bengali is the 7th most popular language in the world and is mainly spoken in the Eastern part of India and in Bangladesh. It is derived from Indo-European languages of the 10th century. Bengali script is used to express the language in written form. Like other Indic scripts the Bengali script has its root in Brahmi script.

Bengali signatures consist of readable text and usually include two parts, the first part refers to the name and the second part refers to the surname of a person. An example can be seen at Figure 3. There is a third word, only when the middle name is included.

To define the name of the synthetic signer, a statistical language model is used. This model should take into account the basic grammar of the language and several other rules as probability distributions of the occurrence frequency of every character that defines the construction of names in this language.

In this way, the number of Bengali signatures with one word is 46.95%, with two words is 46.95% and with third word is 6.1%. The distribution of the number of letters per word is given at Table 1. This data have been worked out from a Bengali Signature Dataset [27].

The probability density function (pdf) of other useful signature morphological parameters are modeled by means of the generalized extreme value (GEV) distribution. GEV combines three simple distributions into a single form, allowing a continuous range of possible shapes. As a consequence, the GEV leads to "let the data decide" which distribution is most appropriate. The GEV is defined as:

$$f(x|\xi,\mu,\sigma) = \frac{1}{\sigma} t(x)^{\xi+1} e^{-t(x)} \quad (1)$$

TABLE 1.
DISTRIBUTION OF WORDS AND LETTERS IN THE SIGNATURE

| Database | Word | \multicolumn{6}{c}{Number of letters per word} |
|---|---|---|---|---|---|---|---|
| | | 1 | 2 | 3 | 4 | 5 | 6 |
| Bengali | 1st | 0.00% | 12.0% | 51.0% | 30.0% | 7.00% | 0.00% |
| | 2nd | 4.00% | 48.0% | 43.0% | 4.00% | 1.00% | 0.00% |
| | 3rd | 0.00% | 69.2% | 23.0% | 0.00% | 0.00% | 7.80% |
| Devanagari | 1st | 0.00% | 17.0% | 53.0% | 20.0% | 10.0% | 0.00% |
| | 2nd | 0.00% | 17.0% | 54.0% | 19.0% | 10.0% | 0.00% |
| | 3rd | 12.50% | 37.5% | 37.5% | 12.5% | 0.00% | 0.00% |

TABLE 2.
VALUES OF THE GEV $\mu, \sigma$ and $\xi$ PARAMETERS FOR THE DIFFERENT DATABASES AND MORPHOLOGY CHARACTERISTICS

| Parameter | Database | $\xi$ | $\sigma$ | $\mu$ | min | max |
|---|---|---|---|---|---|---|
| Total number of letters | Bengali | -0.03 | 1.03 | 5.59 | 4.00 | 12.00 |
| | Devanagari | -0.12 | 1.22 | 5.51 | 3.00 | 11.00 |
| Slant (degrees) | Bengali | -0.26 | 12.32 | -4.75 | -28.50 | 33.69 |
| | Devanagari | -0.16 | 8.25 | -4.39 | -22.31 | 32.47 |
| Skew (degrees) | Bengali | -0.17 | 2.03 | -0.94 | -5.06 | 6.12 |
| | Devanagari | -0.35 | 2.55 | -0.64 | -6.84 | 5.60 |
| Text width (mm) | Bengali | -0.12 | 30.54 | 148.9 | 98.82 | 251.8 |
| | Devanagari | -0.01 | 24.88 | 129.5 | 82.96 | 231.7 |
| Text height (mm) | Bengali | -0.06 | 6.55 | 31.11 | 19.07 | 65.50 |
| | Devanagari | -0.23 | 4.96 | 28.27 | 16.59 | 43.34 |
| Velocity | Bengali | -0.07 | 0.65 | 2.63 | 2.00 | 5.00 |
| | Devanagari | -0.13 | 0.95 | 3.12 | 2.00 | 6.00 |

where

$$t(x) = \begin{cases} (1 + \xi(\frac{x-\mu}{\sigma}))^{-1/\xi} & if\ \xi \neq 0 \\ e^{-(x-\mu)/\sigma} & if\ \xi = 0 \end{cases} \quad (2)$$

being $\mu$ the location parameter, $\sigma$ the scale parameter and $\xi$ the shape parameter.

The values for GEV distribution parameters of signature morphological characteristics such as slant, skew, signature size are given in Table 2. Minimum and maximum values are established for those distributions, as can be seen in Table 2. Signature dynamic distributions such as the velocity of writing the signatures have also been taken into account. Although signatures in Western script exhibit characteristics like skew and slant, such traits are generally much lower in Bengali signatures.

Usually, a name consists of a concatenation of vowels and consonants. In Bengali script there is no concept of an upper case or a lower case letter. All characters in a word are of a similar size, including the starting character. We have considered the most frequently used vowels and consonants in contemporary Bengali script, specifically 12 vowels and 40 consonants.

For Bengali script, the language model turned out to be more complex than English counterpart: this is because (a) there are restrictions on presence/absence of certain vowels before or after the consonants; (b) the shape of the vowels can be changed and this gives rise to vowels in modified form; (c) a modified shape of a vowel may come in two parts and often changes their position with respect to the consonants. Specifically, the following basic rules are followed for automatic generation of a Bengali signature text:
1) A vowel will be followed by a consonant with a probability of 0.97 whereas a consonant will be followed by a vowel with a probability of 0.7.
2) There are certain vowels which are restricted from being the last letter of a word. These are এ, ঐ.

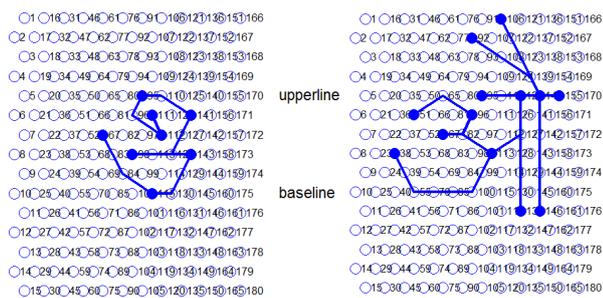

Figure 4. Engram of a Bengali character (left) and Devanagari one (right).

3) A vowel, when it appears after a consonant, changes its shape. For the vowels আ, ী, the vowel modifiers are placed after the consonant. For the vowels ই, এ, ঐ, the vowel modifiers are placed before the consonant. For the vowels উ, ঊ, ঋ, the vowel modifiers are placed beneath the consonant.
4) There are two particular vowels ও and ঔ which, when they appear after a consonant, are split into two parts: one remains in the same position whereas the second part is inserted just before the preceding consonant character.
5) A vowel, when it appears at the first position of a word or after another vowel, keeps its original shape; otherwise it follows the above rules.

Another important aspect of a Bengali signature is that the individual characters generally become connected to each other at the headline region whereas in English signatures, characters mostly become connected in the lower part. However, the presence or absence of character connections are generally random for any particular signature.

*B. Devanagari signature characteristics*

Hindi is the language followed here for Devanagari script. As a language, it stands 4th among all the languages spoken in the world. It is mainly spoken in the Northern part of India and is regarded as the most spoken language of India. Devanagari, again, has its root in the Brahmic scripts.

Signatures written in Devanagari script have little resemblance to Western ones, but are somewhat closer to their Bengali counterpart because of the similar origin of the scripts. Devanagari signatures mainly consist of readable texts. 47.08% of them contain one word, 47.08% two words and the remainder 3.84% three words. The distribution of the number of letters per word worked out from a Hindi Signature Dataset [27] is given in Table 1. Other useful signature morphological parameters, modelled by means of GEV distribution are given in Table 2.

To synthesize the signature name in Devanagari, 11 vowels and 36 consonants are considered. Bearing in mind the frequencies of occurrence of each character, the following basic rules are used:
1) A vowel will be followed by a vowel with a probability of 1% whereas a consonant is followed by a consonant with a probability of 30%.
2) A vowel, while appearing after a consonant, changes its shape. For the vowels आ, ई, ओ, औ the vowel modifiers are placed after the consonant; for इ it is placed before the consonant. For the three vowels उ, ऊ, ऋ the vowel modifiers are placed below the consonant and for the other two vowels ए and ऐ the modifiers are placed on top of the consonant.
3) In general, if the vowel comes at the beginning of the word, it keeps its shape, and if it appears somewhere else, it changes to the modified shape.
4) The first vowel अ keeps its shape at the beginning of a word, but is absent in any other position.

Another important aspect of Devanagari signatures is that the individual characters generally become connected to each other at the upper line region via the matra (see example in Figure 3). There are few vowel modifiers like the modified vowels which finish at either the top of the consonant or the bottom of the consonant. For these, the corresponding character is normally not connected to the following character, but always connected to the previous one. The character connections, i.e. the presence or absence of them, are generally random for a particular signature. The character connection can again be divided into two subtypes: (a) where each subsequent character is connected along the matra (general aspect, not every character needs to be connected) and (b) where the matra is made after a part (greater than a single letter) or the whole of a word.

III. STATIC SIGNATURE DEVELOPMENT

This section proposes a means to generate the signature engram that imitates the cognitive action plan of motor equivalence theory. This is not a simulation of the neurological processes but a procedure inspired by those concepts. Motor equivalence theory suggests that actions are encoded in the central nervous system as a plan. The cortex seems to encode the information about position (i.e. place, distance and direction) in hexagonal mesh whose key unit is the grid cell[28]. Inspired by this idea [19][20] proposed a grid, which spans the signing surface, as the signature engram of a sequence of grid nodes.

The signature engram is described in two steps: the letter engram and the pen-up engram.

*A. Letter engram*

The text engram is built up as a sequence of grid nodes through a tessellation that is calculated by concatenating letter engrams. In English, each letter engram was defined in a hexagonal letter grid of 7 rows and 5 columns [19][20]. But both Bengali and Devanagari characters are more complex and curved than English ones. Hence, a denser mesh is required to define accurately the letter engrams following our earlier proposal [29] for offline Bengali signatures.

Experimentally, a common letter grid for both Bengali and Devanagari scripts has been defined which consists of 180 nodes distributed in 15 rows and 12 columns. The baseline is the 10th row whilst the upper line is assigned to the 5th row (see Figure 4). The rows 11 to 15 are used to accommodate the shape of those characters with strokes in the region below the underline but they are also required to accommodate the modified shape of the vowels. This distribution is also adequate for representing the small circular shapes observed in several Bengali and Devanagari characters.

The distance between columns and rows of the grid is generally different for different writers but constant for each writer. This helps to define the writing style, including the personal letter shape. The letter grid nodes are labelled with a number and each letter engram is defined as a sequence of grid nodes. For instance, the Bengali 'ও' letter engram is defined as the sequence of grid nodes: {111, 112, 96, 95, 110, 141, 127, 97, 127, 143, 130, 115, 84, 67} where the speed minima are at components {2, 4, 6, 8, 10, 12} as can be seen at Figure 4. We refer to speed minima as the minima points in the handwriting speed profile. These minima points set the transition between the strokes.

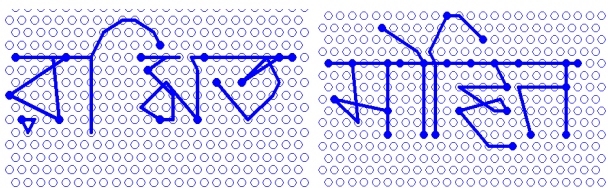

Figure 5. Engram of both Bengali (left) and Devanagari (right) names.

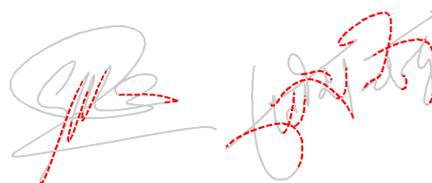

Figure 6. Western and Devanagari signatures. Dashed red lines show the penup trajectory

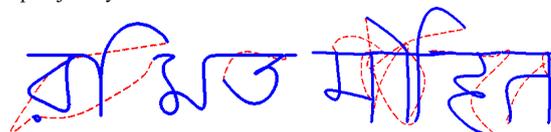

Figure 7. Synthetic trajectory for Bengali (left) and Devanagari (right) signatures

Any particular character can be written in a slightly different shape even by the same writer. Taking this into account, multiple variants of one particular character have been stored. The letter engram definition includes a stroke division. The strokes can be obtained as the nodes where the pen velocity is a minimum in the velocity profile. In this way, the strokes of each letter of both alphabets were defined by the inspection of recorded digital tablet samples and examining minima in the velocity profile. The stroke division nodes can be seen at Figure 4 and Figure 5 as solid nodes.

The signature engram is obtained by concatenating the letter engrams as in Figure 5. Finally, the tessellation is spanned to allow the definition of grid nodes for the pen-up engrams. A characteristic that has to be added to the signature engram, the matra, has received special attention as it can be written after each character or after several characters (the usual case).

*B. Pen-up engrams*

Once the text engrams are defined, it is necessary to define the pen-up engrams that link disconnected handwritten components which appears when the pen is lifted between two consecutive characters. The pen-up engram was modeled in [19][20] by dividing its trajectory into three zones: the start or source area defined as a circle around the pen lift, the intermediate area defined as a tube going from the pen lifting up to pen descent and the sink or ending area defined as a circle around the pen descent. A number of nodes are randomly selected in each area and the pen-up engram is defined by linking them. The radius of the source and ending circles, along with the radius of the tube, define the straightness of the pen-up trajectory and the number of nodes define its randomness. For a Western script, the radius of the circle is set to $d/10$, $d$ being the distance between the beginning and ending of the pen-up, and just one node is selected in each zone.

In general, Western writers display very straight pen-up trajectories whereas Bengali and Devanagari writers, in our databases, show pen-ups with more rounded trajectories which follow the curved handwriting style of Bengali writers, as can be seen in the example of Figure 6.

Moreover, from analyzing the speed profiles, it seems that Western writers slow down the pen speed at the beginning and end of every pen-down while the Indian writers slow down the pen at the end of every pen-down but they do not reduce the pen speed during the transition from pen-up to pen-down. This produces a sort of harpoon shape at the beginning of the majority of the pen-downs. These "harpoons" cannot be observed in Devanagari. It appears that this different handwriting style stems from the learning procedure at primary school. In fact, in our experiments, native Bengali writers use the same method for writing both Bengali and English,

This behavior has been confirmed with the Fourier Transform of the $x$ and $y$ sequences of the signature trajectory. It was supposed that their spectral bandwidth would be greater for Bengali and Devanagari than for Western scripts because the Indian characters contain more corners than Western ones. However, the opposite is the case: the bandwidth of the Indian writers is narrower than for the Western writers because of the rounding way in which native Indians write. It seems that Indian writers generate more circular/curve shapes in their handwriting so they can write more complex characters when compared with Western writers, who produce less complex characters in the same time period.

Therefore, the pen-up engram model is changed to produce curved pen-up trajectories. This is conducted by selecting one or two points only in the intermediate area whose radius is heuristically widened to $d/3$ with a minimum value of $d/10$. The effect of this procedure can be seen in the synthetic pen up trajectories in Figure 7. Additionally, the stroke limits at the beginning of the pen-downs are removed, thus integrating both the pen-up ending and pen-down starting points in the same speed cycle.

*C. Signature trajectory: motor control*

Once the signature engram is defined, an inverse model for motor control is applied to obtain a realistic human signature ballistic trajectory[23]. In the Western signature synthesizer [19][20], the signature ballistic trajectory is worked out by using a multilevel motor scheme based on inertial moving average filters that emulate the inertia of different muscles used for handwriting. In summary, it obtains the signature trajectory by filtering the engram with three inertial filters which are assigned heuristically to the action of fingers, forearm and wrist: the finger inertial filter is applied to the text engram, the forearm filter is applied to the flourish engram and the wrist filter is applied to the whole signature.

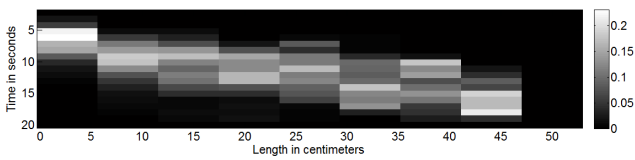

Figure 8. Joint probability density function of the signature time and signature length.

For both Indian scripts, a similar scheme is applied, but, as Devanagari and Bengali script, in general, consist of shorter strokes than Western signatures, the wrist filter is reduced to the minimum and the finger and forearm filters turns out to be relevant for defining the signature trajectory from the signature engram.

Another characteristic that differs from Western to Indian languages is the significance of the stroke contact points. To preserve the position of the contact point is essential for keeping the legibility of the character. In turn, these contact points are automatically detected and the filtered trajectory is constrained to pass through these points.

Bearing in mind these considerations, the signature trajectory is obtained by linking the engram nodes by the Bresenham's line algorithm. The inertial filters are then applied to the line. The finger control motor filter is applied to the shorter strokes, stopping at every stroke limit. The forearm control filter is applied to the larger strokes and the wrist control motor filter is applied to the whole engram to obtain the signature trajectory. The inertia filters are based on Kaiser filters with symmetric finite impulse response $h(n)$ defined as:

$$h(n) = \begin{cases} I_0(\pi\beta\sqrt{1-(2n/N^n-1)^2}) & 0 \le n \le N \\ 0 & otherwise \end{cases} \quad (3)$$

where $N$ corresponds to the filter length and $\beta$ is experimentally set to 0.1. The value of $N$ is related to the signing velocity $v$ which is randomly obtained following its probability density function as defined in Table 2. The finger, forearm and wrist filter lengths are worked out as $N_f = L_f v^2$, $N_a = L_a v^2$ and $N_w = L_w v^2$ respectively. The inertia variables $L_f$, $L_a$ and $L_w$ define the synthetic user inertia and are randomly set from 3 to 4 times the distance between the grid nodes. These values are kept constant for each user. Due to the smaller bandwidth of the Indian writing, the inertial filters are longer than in the case of the Western scripts.

A drawback of this procedure is that the trajectory of straight vertical and horizontal strokes appears unnaturally straight. This problem is reduced by converting the long straight lines into triangles. The triangle height is a constant for each writer and in the range $[0, d/10]$, where $d$ is the length of the straight stroke. Figure 7 shows the resulting trajectory obtained from the above handwritten engrams.

Finally, a realistic offline signature image is generated by applying the ink deposition model proposed in [18] to the trajectory.

## IV. SIGNATURE DYNAMICS

This section is devoted to sampling the continuous signature trajectory so as to provide the synthetic online signature from the above trajectory in a unified synthesis framework. The major problem for the sampling is to obtain a human like speed profile $\bar{v}(t)$, which the kinematic theory of the rapid movements claims to be a linear combination of lognormals as follows:

$$\bar{v}(t) = \sum_{j=1}^{M} \bar{v}_j(t; D_j, \tau_j, \mu_j, \sigma_j^2) \quad (4)$$

where the speed profile of each stroke $v_j(t)$, is defined as:

$$v_j(t; \tau_j, \mu_j, \sigma_j^2) = \frac{D_j}{\sigma_j\sqrt{2\pi}(t-\tau_j)} exp\left\{-\frac{[ln(t-\tau_j)-\mu_j]^2}{2\sigma_j^2}\right\} \quad (5)$$

where $t$ is the time, $\tau_j$ is the time of stroke occurrence, $D_j$ is the amplitude of each stroke, $\tau_j$ is the stroke time delay on a logarithmic time scale and $\sigma_j$ is the stroke response time, also on a logarithmic time scale. Consequently, two tasks have to be performed, the first to synthesize the speed profile and the second to sample the signature trajectory for fitting the given speed profile.

### A. Velocity profile synthesis

Let us suppose a single stroke velocity profile is given by $v_j(t)$. The values of $D_j$, $\mu_j$ and $\sigma_j^2$ are set from the following two hypotheses: first, the margins for natural human handwriting given in [29] and second, experimentally it is observed that most of the lognormals are centered, i.e. the lognormal peak approaches the stroke center from the left. Therefore, the skewness is close to zero but positive and the kurtosis is around 3. This conclusion was drawn from 300 different signatures selected from each respective dataset. Their lognormal decomposition was worked out with the ScriptStudio program[29]. Thus the values $D_j$, $\mu_j$ and $\sigma_j^2$ are randomly obtained for each stroke within the margins specified in [29] and iteratively refined to fit the skewness in the range $[0 – 0.3]$ and Kurtosis [3-3.5].

The occurrence time for each stroke $\tau_j$ is computed from the so-called Central Pattern Generators (CPG) that produce rhythmic patterned outputs, without sensory feedback, to activate different motor pools[30]. This can be observed in the clearly periodic pattern of the handwriting speed.

Therefore, if the stroke generation is assimilated into the CPG step cycle, the duration of each stroke is very similar to any other. This has been verified with the above mentioned signature samples where each signature is divided into strokes to calculate the dispersion of the strokes' duration. The dispersion is defined as the standard deviation divided by the mean. The distribution of the dispersion can be approximated by a Gaussian distribution $N(0.32,0.06)$. Consequently, the duration assigned to each stroke of the synthetic signatures is computed by (i) dividing the whole signature duration by the number of strokes and (ii) multiply by a random sequence to fit the dispersion to the time sequence assigned to each stroke.

The signature duration is obtained randomly from the joint probability density function that relates signature time and length, which is shown at Figure 8. The length is known from the trajectory defined above.

### B. Lognormal sampling of the trajectory

As the trajectory has been defined in space, the time $t(s)$ of every pixel on the trajectory is required for spatial sampling. This is calculated as follows. Let us assume that a single

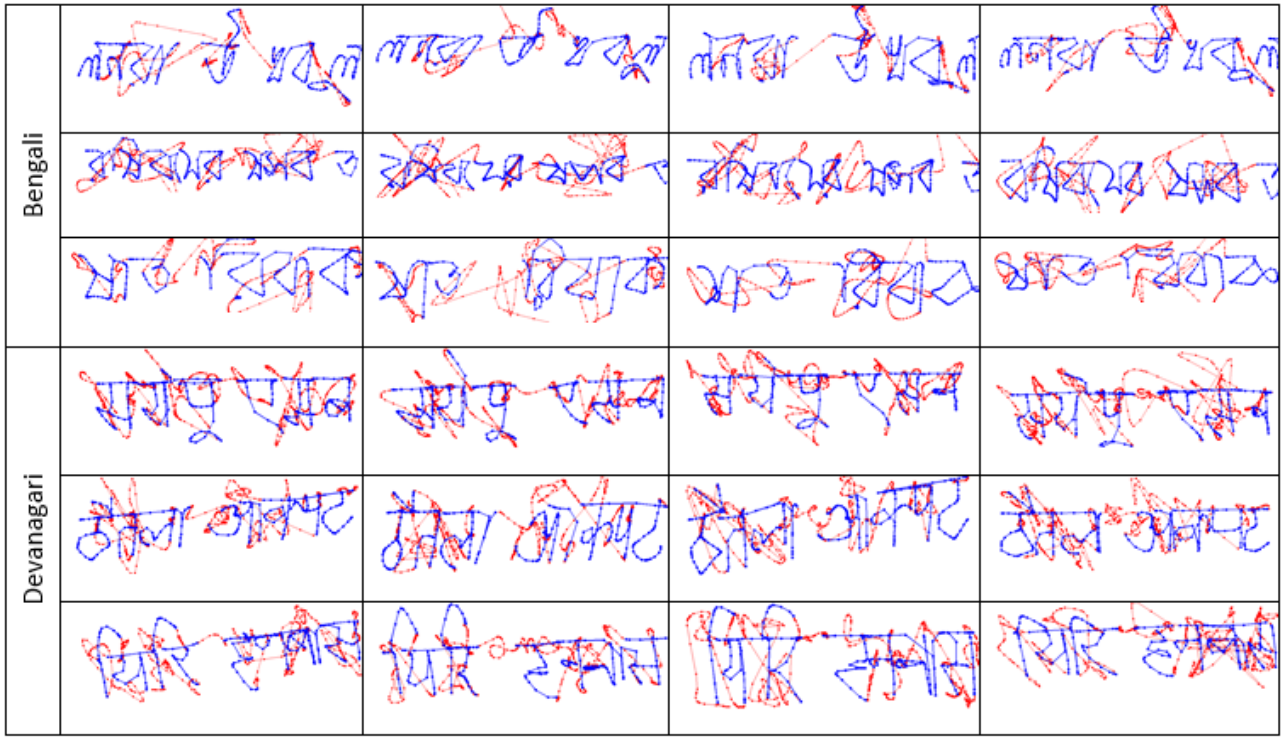

Figure 9. Examples of generated intra and inter personal variability for three Bengali and three Devanagari signatures.

stroke's velocity is given by $v_j(t)$ in Eq.(3). The distance traversed at time $t$ is then obtained as:

$$s(t) = \int_{-\infty}^{\infty} v_j(t)dt = = \frac{D_j}{2}\left(1 + \text{erf}\left(\frac{\ln(t-\tau_j) - \mu_j}{\sqrt{2}\sigma_j}\right)\right) \quad (6)$$

which is the lognormal cumulative function. Solving for $t$ in this equation, we get the time in terms of the distance as:

$$t(s) = \exp\{\sqrt{2}\sigma_j \text{erf}^{-1}(2s/D_j - 1) + \mu_j\} \quad (7)$$

where the values of $D_j$, $\mu_j$, $\sigma_j$ and $\tau_j$ have been computed in the above section. The time at every pixel in the signature trajectory is known from Eq. 7. Then the accumulated time along the signature trajectory is calculated and the trajectory is sampled by selecting the pixels for which the time is closer to multiples of $1/f_m$, where $f_m$ is the sample frequency. Figure 9 shows the realism of the synthetic dynamic information.

V. GENERATION OF MULTIPLE IDENTITIES AND SAMPLES

In this section we describe the flexibility of the synthesizer to adjust both inter and intra personal variability. The evaluation in the following section consists of adjusting these variabilities so that they are similar to native Bengali and Devanagari handwriting.

A. *Generation of multiple identities: inter personal variability*

In the synthesizer, inter personal variability is introduced at morphological, cognitive and motorial levels:
- *Morphological level:* As Indian signatures consist mainly of the signer name and surname, the morphological variability is achieved by randomly generating the signer's full name as per the statistics defined in section 2. Additional morphological parameters considered are: signature skew, letter slant, average space between letters and average space between words.
- *Cognitive level*: This variability is introduced by adjusting the distance between rows and columns of the signer grid. This distance is randomly defined for each writer between a maximum and minimum dimension heuristically selected. It defines the text height and width and provides different letter styles for each signer.
- *Motorial level*: The signer identity significantly depends on his or her average writing velocity which is related to inertial filter lengths, namely the $L_f$, $L_a$ and $L_w$ parameters defined in section 3.3. The stroke duration dispersion (also called jitter), the lognormal skewness and kurtosis and the straightness of the lines are also defined for each user at this stage.

The signer stability parameter is also defined at this point because some signers are more stable than others. There are two signer stability parameters: the first refers to static features, namely the morphological and cognitive ones, the second refers to dynamic features, such as the speed and motor filters. At this point, the obtained signature is called the *master signature* of the synthetic signer and the related variables are retained for the rest of its genuine duplicates.

Several Bengali and Devanagari examples of online signatures synthetically generated via the proposed procedure are shown at Figure 9. The blue dots indicate the pen-downs whereas the red dots refer to pen-ups. It is worth highlighting that the synthesizer is capable of generating a wide range of signatures.

*B. Multiple sample generation: intra personal variability*

Every new sample of a genuine signature is different from the previous one because of slight changes in posture, writing tools, emotional condition and so on. So, once the master signature has been generated, the intra personal variability is applied to this master signature by random modification of the master signature parameters.

Parameters such as signature skew, letter slant, average space between letters and average space between words, are varied as follows: let $p$ be the parameter value and $mp$ its range, i.e. the maximum value minus the minimum. Let $s$ be the stability value for static parameters. The value $p$ is computed for every sample generated for the same signer as $p + u$, $u$ being a random variable $N(0, s*mp)$. If the modified value exceeds the parameter range, it is set to the minimum or maximum value accordingly.

As Indian scripts consist of multiple short strokes, the order of them is changed frequently, thus morphological variability has been also taken into account as follows: the intra personal changes focus on the matra, the vowel modifiers and the character morphology. Instead of writing a matra for each character, the writers tend to elongate them over different characters. Thus, the number of matras and their length are a source of morphological intra-personal variability. In a similar way, the character style is changed depending on whether it is in the scope of the matra. Likewise, some characters and some vowel modifiers are changed with the above mentioned probability depending on how the previous and the following character take shape.

At cognitive level, the intra personal variability is dealt with as follows:
1. The grid nodes change their position randomly inside a circumference centered on the particular grid node. The radius of the circumference is another parameter of the signer. This radius is reduced for the grid nodes that belong to the baseline and the upper line (see Figure 3) because the Indian way of handwriting is more stable in these grid nodes.
2. A sinusoidal transform is applied to the signature engram nodes as in [8]. This deformation aims to replicate approximately the effect of synchronism variability among agonist and antagonist muscles.
3. An affine distortion is applied by changing the relative $x$ and $y$ scales. The scale for each segment is randomly selected between 1-$s$ and 1+$s$ with $s$=0.1.
4. The signature engram nodes for each pen-down are displaced toward the right to a distance randomly chosen between zero and twice the grid distance between nodes, and vertically to a distance randomly chosen between zero and half the grid distance between nodes.

At motor level the parameters are modified in the same way as the morphological ones. In Figure 9 we show several examples of generated intra personal variability for some Bengali and Devanagari signatures.

VI. EXPERIMENTS

The objective of these experiments is to evaluate the similarity between synthetic and real signatures. With this aim, a dual static and dynamic signature database has been recorded for Bengali and Devanagari scripts. Native writers have been involved in this task. Two kinds of experiment have been performed: performance based and perception based. The performance experiments compare the Equal Error Rate (EER) and Detection Error Trade-off (DET) curves for the synthetic and real databases using different classifiers: two online and two offline. A DET plot represents the false rejection rate (FRR) curve vs. false acceptance rate (FAR) curve. Then, the ERR is the point in the DET plot at which both curves (FRR and FAR) are equal.

The perceptual experiment is carried out by surveys to estimate the human ability to distinguish between the synthetic and real signatures by naturally viewing those signatures.

*A. Third party offline Bengali and Devanagari databases*

Two third party, available databases have been found in the literature. The first one, an offline Bengali signature database [27] which contains 100 users with 24 genuine signatures and 30 forgeries per user. For each contributor, all genuine specimens were collected in a single day's writing session. In order to produce the forgeries, the imitators were allowed to practice their forgeries as long as they wished from static images of genuine specimens.

The second database, an off-line Hindi dataset [27], is similar to the Bengali one. It consists of 24 genuine signatures and 30 forgeries per user.

*B. Description of OnOffSigBengali-75 and OnOffSigHindi-75 corpuses*

An additional contribution of this work is the development of OnOffSigBengali-75 and OnOffSigHindi-75 corpuses, which contain static and dynamic signatures simultaneously captured for both scripts. The main motivation was to alleviate the setback of the above third party signatures which are useful for comparing synthetic and real databases in offline procedures but not in the online mode.

OnOffSigBengali-75 and OnOffSigHindi-75 corpuses were collected over a WACOM Intuos 3 Tablet with an ink pen, so the online and offline versions of each signatures are available. The inked signatures were scanned at 600 dpi to obtain the offline signature. The image background was removed following the procedure suggested in [31]. The images were saved in .jpg format. For the online version, the sampling frequency is 100 samples per second and the resolution 2540 dpi. Due to some temporal irregularities, the samples were spline interpolated to assure a uniform sampling frequency.

Regarding the size of databases, they consist of signatures from 75 native Bengali and 75 native Devanagari writers. Each writer produced 24 signatures in two sessions, 12 in each session. The height and width of the form signing boxes were 3 cm and 12 cm respectively.

These four novel databases are publicly available for research purposes and they can be downloaded from http://www.gpds.ulpgc.es/.

*C. Perceptual experiments*

We conducted two visual Turing tests to investigate the generator's ability, according to human judges, to produce humanlike signatures. The two visual Turing test were carried out independently for Bengali and Devanagari scripts and performed by non-forensic volunteers and Forensic Handwriting Experts (FHEs).

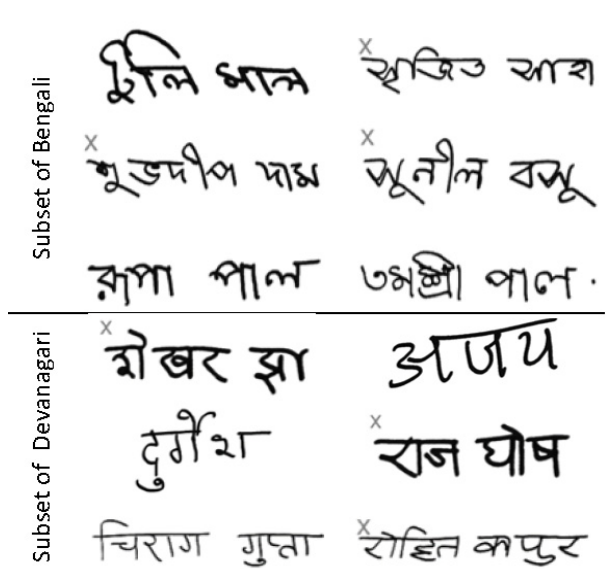

Figure 10. Subset of signatures used in the visual Turing Tests to validate the appearance of synthetic samples. Synthetic signatures are marked with a cross.

In a similar way to [13][19], the respondents were asked to score between 0 (very sure that it is synthetic) and 10 (very sure that it has been written by a human) each signature after a careful inspection.

The visual Turing tests combined real signatures randomly selected from our own recorded datasets with synthetic specimens. Exactly 110 non-forensic volunteers and 8 FHEs participated in these tests. They were of Indian origin and had a good knowledge of each script. In total, volunteers made more than 10,000 decisions.

For a fair comparison, the signatures were segmented to be shown over the same white background and the ink of all the signatures were substituted by the ink deposition model [18]. A subset of the survey sheets used in these experiments is shown in Figure 10, where the synthetic specimens are marked with a cross.

The perceptual realism of the synthetic generator is measured by calculating two kinds of errors: the False Synthetic Rate (FSR: rate of real signature with a score less than 5) and the False Real Rate (FRR: rate of synthetic signature with a score greater than 5). The Average Classification Error (ACE) is calculated as ACE= (FSR + FRR)/2. The results of both surveys are shown in Table 3, where the number of effective decisions that we registered from the volunteers are given.

A complete confusion is achieved at 50% of Error Rate. The visual Turin Test made by non-forensic volunteers confirm the high confusion to discern whether our synthetic specimens are real or not. Therefore, judging by the Turing tests, the results suggest that perceptually the synthetic signature generator based on the motor equivalence approach is able to produce signatures that native human, non-expert examiners accept visually as real.

TABLE 3.
VISUAL TURING TEST RESULTS

| BENGALI | | Non-forensic | FHEs |
|---|---|---|---|
| Number of participants | | 70 | 4 |
| Number of answers | | 7360 | 320 |
| Error Rates (%) | FSR | 30.90% | 35.50% |
| | FRR | 53.26% | 11.88% |
| | ACE | 42.08% | 22.19% |
| Averaged scores | Real | 6.50 | 5.19 |
| | Synthetic | 5.30 | 1.58 |
| Averaged time per signature | | 7.54 sec. | 15.38 sec. |
| DEVANAGARI | | Non-forensic | FHEs |
| Number of participants | | 40 | 4 |
| Number of answers | | 3776 | 128 |
| Error Rates (%) | FSR | 44.44% | 29.69% |
| | FRR | 42.11% | 3.13% |
| | ACE | 43.27% | 16.41% |
| Averaged scores | Real | 8.74 | 7.04 |
| | Synthetic | 4.85 | 0.70 |
| Averaged time per signature | | 7.49 sec. | 15.84 sec. |

TABLE 4.
EER COMPARING THE PERFORMANCE OF OFFLINE AND ONLINE SYNTHETIC DATASETS WITH REAL ONES

| Training | Classifier | Bengali | | | Devanagari | | |
|---|---|---|---|---|---|---|---|
| | | Third Party | In House | Synthetic | Third Party | In House | Synthetic |
| 2 | HMM | 6,03% | 5,54% | 6,24% | 5,32% | 6,71% | 5,73% |
| | SVM | 4,32% | 0,84% | 2,73% | 2,85% | 1,37% | 2,03% |
| | DTW | NA | 0,41% | 1,66% | NA | 0,41% | 1,02% |
| | Man | NA | 8,19% | 12,6% | NA | 9,77% | 12,2% |
| 5 | HMM | 4,08% | 3,18% | 3,06% | 3,37% | 4,43% | 2,68% |
| | SVM | 1,97% | 0,23% | 0,67% | 1,56% | 0,56% | 0,47% |
| | DTW | NA | 0,27% | 0,47% | NA | 0,31% | 0,49% |
| | Man | NA | 3,41% | 6,50% | NA | 4,16% | 5,79% |
| 8 | HMM | 3,17% | 2,46% | 2,38% | 2,77% | 3,58% | 1,75% |
| | SVM | 1,32% | 0,13% | 0,34% | 1,37% | 0,38% | 0,24% |
| | DTW | NA | 0,33% | 0,25% | NA | 0,34% | 0,36% |
| | Man | NA | 2,7% | 5,44% | NA | 3,59% | 4,7% |
| 10 | HMM | 2,5% | 2,34% | 1,71% | 2,77% | 3,22% | 1,34% |
| | SVM | 1,12% | 0,13% | 0,25% | 1,33% | 0,31% | 0,14% |
| | DTW | NA | 0,28% | 0,23% | NA | 0,29% | 0,38% |
| | Man | NA | 2,57% | 5,11% | NA | 3,11% | 4,41% |

NA stands for 'Not Apply' as Third Party dataset does not contain online signatures

Regarding the tests performed by the Forensic Handwriting Experts, the results obtained highlight the level of agreement attained by them. Despite carrying out their test independently, the Pearson correlation coefficient between their answers of Bengali and Devanagari tests are 0.49 and 0.90 respectively. It confirms a statistical relationship between their judgements.

Compared with non-forensic volunteers, the FHEs detected correctly the synthetic signatures (lower FRR). However, the FHEs and non-forensic volunteers show a similar degree of confusion in identifying the real signatures (FRR>30). In addition to spending more time inspecting each signature, it can be noted that the FHEs are more skilled than non-forensic volunteers in detecting synthetic signatures.

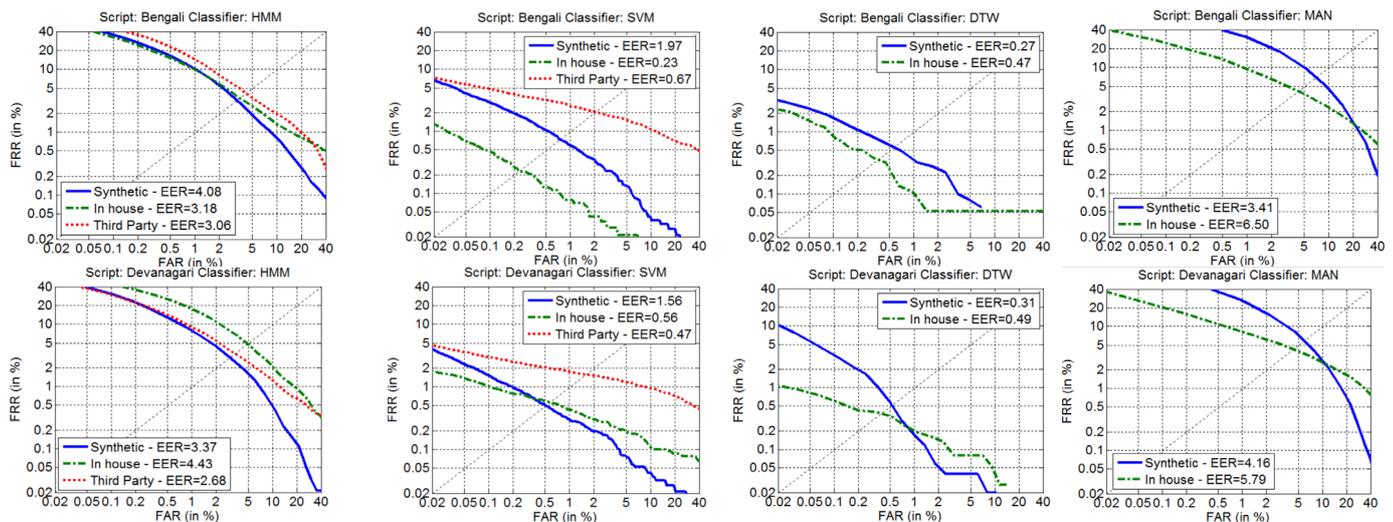

Figure 11. DET plot comparison for all systems using the third party, in house and synthetic database for both scripts training with 5 signatures

### D. Performance experimental protocol

For the performance experiments, four Automatic Signature Verifiers (ASVs) have been chosen: two offline and two online. We chose these four ASVs on the basis of conceptually different features because they are expected to cover a wide range of signature properties. The classifiers are, for static signatures:
1. Hidden Markov Model (HMM) based verifier: geometric features and HMM as classifier. The static signature is parameterized in Cartesian and polar coordinates and combined at score level. A multi-observation discrete left to right HMM is chosen to model each signer's [32].
2. Support Vector Machine (SVM) based verifier: in this case, the Local Binary Pattern (LBP) operator has been used for static signature parameterization. A Least Squares Support Vector Machine (LS-SVM) with RBF kernel has been used as classifier. The procedure is described in [31].

For dynamic signature verification, we use:
1. A Dynamic Time Warping (DTW) based verifier: in this case, the dynamic sequences of one enrolled signature and a questioned specimen are compared using a DTW algorithm with Euclidean Distance [33].
2. A Manhattan distance based verifier: the feature vector consist on two histograms: one with absolute frequencies and other with relative frequencies. The histograms of questioned and unquestioned signatures are compared with Manhanattan distance [34].

All the verifiers are trained by following the same well-established experimental protocol in which the training set consists of $T$ randomly selected genuine signatures. The remaining genuine signatures are used for testing the false rejection rate. In all cases, the false acceptance rate has been obtained with the genuine test samples from all the remaining users. All the experiments are repeated 10 times. For instance, if $T=5$ training samples, the number of tests made for establishing the false rejection rate with the in house dataset (75 users) for each script are $(24-5) \times 75 \times 10 = 14250$ whereas for the false acceptance rate the number of test samples are $(24-5) \times 75 \times 74 \times 10 = 1054500$. All of these experiments are conducted for $T=2$, $T=5$, $T=8$ and $T=10$.

### E. Performance Experimental Results

To evaluate the similarity between the real and synthetic databases, a synthetic database was generated for each script. They consist of 100 synthetic signatures with 24 samples per signature. For each signature the offline and online versions were synthesized.

The setup of the synthesizer to generate the databases was conducted in four steps:
1) Introduce the parameters of the morphology of the real dataset into the synthesizer.
2) Form the variability parameters for genuine static signature generation to approach the performance for the offline random impostor experiment.
3) Form the parameters for genuine dynamic signature generation to approach the performance for online random impostor experiment.
4) If the differences between the real and synthetic performances are larger than expected in the last step, the procedure is iteratively repeated looking for the minimum square error between the eight EERs of the real and synthetic dataset.

This procedure is based on the hypothesis that the parameters used for generating the dynamics of the signature do not affect the performance of the static signature, but the opposite does not apply. This was performed by training with $T=5$ samples. The results of the experiments are shown in Table 4.

As can be seen, all real and synthetic databases perform in the same range and display similar trends in all the cases with $T=2$, $T=5$, $T=8$ and $T=10$ which proves the ability of the synthesizer to generate databases with performances similar to those obtained with real datasets, offline, online and for different classifiers and training samples. The DET curves for the case $T=5$ are provided in Figure 11 which reinforces the conclusions obtained from Table 4.

## VII. Conclusion

This paper proposes a unified theoretical framework for generating both online and static synthetic signatures in two of the most popular Indian scripts: Devanagari and Bengali. The synthesizer is based on the motor equivalence model that emulates the human way of performing actions. The specific way of writing these two Indic scripts generates more complex symbols which are composed of two or more short components with a few strokes.

So this paper proposes significant new developments in the application of the motor equivalence model to handwriting generation in these two Indic scripts. The main improvement of the model relating to the cognitive level relies on the variable density of the grid that defines the signature engram. The Fourier transform is used for adjusting the inertial filter length at motorial level. A procedure to balance the variability between the cognitive and motor system has been proposed to achieve humanlike variability when Indian scripts are generated.

To evaluate the ability of the proposed synthesizer, dual offline and online Bengali (OnOffSigBengali-75) and Devanagari (OnOffSigHindi-75) signature corpuses have been produced as no such datasets were previously available. The performance of both real and synthetic databases are similar. A perceptual experiment to evaluate the likeness of real and synthetic database, obtained over 42% confusion and this indicates the realism of the synthesized signatures.

The feedback of these findings to the Western Signature Synthesizer is expected to improve its "humanity", i.e. to increase its versatility. Besides, it will allow a future multi-script handwritten signature generator for biometrics and other areas such as health care, forensics, etc. The new datasets are freely available to other researchers in this field.

> REPLACE THIS LINE WITH YOUR PAPER IDENTIFICATION NUMBER (DOUBLE-CLICK HERE TO EDIT) <   12

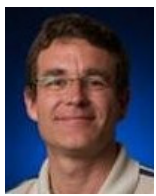
**Miguel A. Ferrer** received the M.Sc. and Ph.D. degrees from the Universidad Politécnica de Madrid, Madrid, Spain, in 1988 and 1994, respectively. He joined the University of Las Palmas de Gran Canaria, Las Palmas, Spain, in 1989, where he is currently a Full Professor. He established the Digital Signal Processing Research Group in 1990. His current research interests include pattern recognition, biometrics, audio quality, and computer vision applications to fisheries and aquaculture.

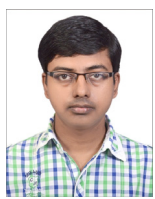
**Sukalpa Chanda** obtained PhD in Computer Science in June 2015 from Gjøvik University College in Norway (currently known as NTNU IN Gjøvik) and is presently a postdoctoral researcher in the Institute of Artificial Intelligence and Cognitive Engineering at University of Groningen, Netherlands. His research interest includes Pattern Recognition, Document Image Analysis, Computational Forensics, etc.

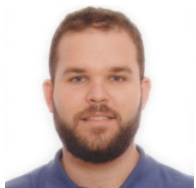
**Moises Diaz** (M'15) received the M.Tech., M.Sc., and Ph.D. degrees in engineering, and the M.Ed. degree in secondary education from La Universidad de Las Palmas de Gran Canaria, Las Palmas, Spain, in 2010, 2011, 2016, and 2013, respectively. He is currently an assistance professor at Universidad del Atlantico Medio, Spain. His current research interests include pattern recognition, document analysis, handwriting recognition, biometrics, computer vision, and intelligent transportation systems.

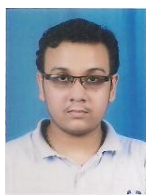
**Chayan Kumar Banerjee** is currently in his final year of his undergraduate program in Mathematics and Computer Science from Chennai Mathematical Institute. He has done academic projects at Chennai Mathematical Institute and Indian Statistical Institute, Kolkata. His areas of interest include Combinatorics, Algorithms and Applied Areas of Computer Science.

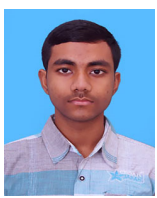
**Anirban Majumdar** is currently pursuing his B.Sc. degree in Chennai Mathematical Institute, India in Mathematics and Computer Science. His areas of interest are Automata Theory, Mathematical Logic, Concurrent Systems and Pattern Recognition. He has been a visiting student in the Indian Statistical Institute.

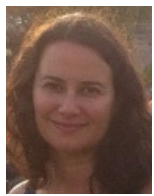
**Cristina Carmona-Duarte** received the Telecommunication Engineering degree in 2002 and de Ph.D. degree in 2012 from Universidad de Las Palmas de Gran Canaria. She has been assistant Professor at Universidad de Las Palmas where she is currently a researcher. Her research areas include high resolution radar, pattern recognition and biometrics.

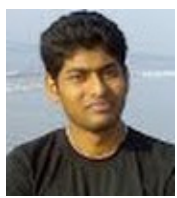
**Parikshit Acharya** obtained a Bachelor of Technology degree and is currently working as one of the project linked personnel in the Computer Vision and Pattern Recognition unit of the Indian Statistical Institute, Kolkata. His research interests lie in the field of document image analysis.

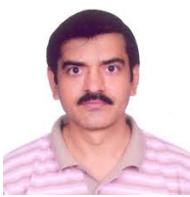
**Umapada Pal** received his Ph.D. in 1997 from Indian Statistical Institute. He did his Post-Doctoral research at INRIA (Institut National de Recherche en Informatique et en Automatique), France. From January 1997, he is a Faculty member of Computer Vision and Pattern Recognition Unit (CVPRU) of the Indian Statistical Institute, Kolkata and at present he is Professor and Head of CVPRU. His fields of research interest include Digital Document processing, Optical Character Recognition, Biometrics, etc.